%% file: main.tex
\documentclass[sigconf,nonacm]{acmart}
\usepackage{amsmath}
\AtBeginDocument{%
  }

\setcopyright{none}

\begin{document}

\title{Hidden in Plain Gaze: Gaze Representations as Privacy Controls for Utility and Re-identification Risk in XR}

\author{Cory Ilo}
\affiliation{%
  \institution{Virginia Tech}
  \city{Blacksburg}
  \state{Virginia}
  \country{USA}}

\author{Brendan David-John}
\affiliation{%
  \institution{Virginia Tech}
  \city{Blacksburg}
  \state{Virginia}
  \country{USA}}

\author{Doug A. Bowman}
\affiliation{%
  \institution{Virginia Tech}
  \city{Blacksburg}
  \state{Virginia}
  \country{USA}}

\renewcommand{\shortauthors}{Ilo et al.}

\begin{abstract}
Intelligent extended reality (XR) systems increasingly use eye and head
tracking to infer user intent, task, and attention, but the same signals can
also reveal biometric identity. We study whether gaze data
representation choice can serve as a lightweight privacy control at feature
extraction, before adding perturbation or formal privacy mechanisms. Using the
egocentric HoloAssist dataset, we compare three gaze representations under
matched model capacity: raw gaze, spatial attention heatmaps, and engineered
eye-movement features. We evaluate each representation on action recognition as
task utility and closed-set user re-identification as privacy leakage.
Representation choice substantially changes the privacy-utility tradeoff.
Engineered features retain roughly $85\%$ of raw gaze's action-recognition
accuracy while reducing re-identification by about an order of magnitude, to
roughly four times the chance rate across $206$ identities. This reduction
attenuates rather than eliminates identity leakage, and the differences across
representations show that abstraction alone does not guarantee privacy.
Engineered features expose interpretable and auditable structure, giving
designers a transparent privacy lever that complements mechanisms such as
differential privacy.
\end{abstract}
\newcommand{\bdj}[1]{{\footnotesize\color{red}[Brendan: #1]}}



\maketitle

\section{Introduction} 
    \input{sections/intro}

\section{Related Work} 
    \input{sections/rw}

\section{Threat Model}
    \input{sections/threatScenario}

\section{Methodology} 
    \input{sections/methods_revised}

\section{Experimental Results}

    \input{sections/results_revised}

\section{Discussion, Limitations, \& Future Work}
    \input{sections/discussion_revised}

\section{Conclusion}
    \input{sections/conclusion_revised}

\begin{acks}
To Robert, for the bagels and explaining CMYK and color spaces.
\end{acks}

\bibliographystyle{ACM-Reference-Format}
\bibliography{InPlainGaze_GBTA}

\end{document}

%% file: sections/intro.tex
Intelligent extended reality (iXR) systems increasingly rely on continuous egocentric sensing to enable context-aware interaction~\cite{AriaGen2}. Eye gaze and head motion, in particular, have emerged as key input streams that support inference of user intent, task state, and attentional focus \cite{David2021, Hu2023}. At the same time, these signals inherently encode biometric information: patterns of eye movement behavior are highly distinctive across individuals and can be used to reliably distinguish users \cite{Lohr2022}. This dual role suggests a central challenge for iXR system design: the same signals that enable adaptive, low-friction interaction may also expose persistent identity information.

We argue that this tension is not solely a property of the sensing modality itself, but is critically shaped by how that data is represented. In particular, identity-related information in gaze is largely carried by fine-grained temporal dynamics, such as velocity profiles, fixation durations, and saccadic patterns, while task-relevant information is often reflected in coarser spatial distributions of attention over the environment \cite{Rigas2018, Hu2023, Bulling2011}. This separation raises a key question: can representation design selectively suppress identity-bearing signals while preserving task-relevant utility?

Existing privacy-preserving approaches offer partial remedies. Formal mechanisms such as Differential Privacy (DP) provide rigorous guarantees but are typically applied at the dataset, training, or model-output level and can degrade downstream task performance \cite{Ibragimov2025, Kundu2026}. Moreover, integrating such mechanisms into real-time iXR pipelines can introduce non-trivial computational overhead and latency, which may be incompatible with the tight responsiveness requirements of interactive systems \cite{Wilson2024}. These constraints motivate an alternative strategy: rather than modifying signals after collection, can the choice of input representation itself act as a lightweight, inference-time mechanism for controlling information exposure?

Prior work provides indirect evidence for this possibility. Spatial aggregation of gaze into heatmap representations collapses fine-grained temporal dynamics while preserving spatial attention patterns relevant for task classification \cite{Duchowski2012,Burch2013}. Gaze-based representations also enable activity recognition without RGB imagery, mitigating bystander exposure risks common in egocentric video systems \cite{Steil2019}. Differential privacy has also been explored at the feature level rather than on raw samples, but only on a smaller, non-immersive VR dataset for a 2D document-type reading task~\cite{Bozkir2021}. However, prior research has largely studied gaze-based activity recognition and gaze biometrics for re-identification in isolation. As a result, it remains unclear how representation choices jointly influence both task utility and identity leakage under a unified experimental setting.

In this work, we address this gap through a controlled empirical study using the HoloAssist dataset \cite{HoloAssist}, a large-scale egocentric benchmark of procedural tasks collected with mixed-reality headsets. We compare multiple gaze representations, including raw gaze vectors, hand-engineered features derived from eye movement events, and spatial heatmaps, pairing each with an architecture appropriate to its structure. For each representation, we evaluate (1) utility, measured as coarse-grained task classification accuracy, and (2) privacy risk, measured via user re-identification performance, a commonly used proxy for identity leakage in behavioral biometric systems \cite{DJohn2022}. While re-identification does not capture all forms of information leakage, it provides a concrete and measurable lower bound on privacy risk. Our experimental design controls for model capacity---the trainable-parameter budget available to each encoder---enabling a more direct assessment of representation-level effects.

Across these representations, the choice of gaze encoding behaves as a
practical privacy lever applied at the point of feature extraction, before any data-level perturbation is introduced. A continuous (window-based) engineered feature representation retains the bulk of action-recognition utility (about $85\%$ of
raw gaze's top-1 accuracy at the action level and $90\%$ at the verb level) while reducing closed-set re-identification by roughly an order of magnitude, from $\approx\!39\times$ chance to $\approx\!3.9\times$ chance across
$N{=}206$ identities. This attenuates identity leakage rather than eliminating
it, as a residual above-chance signal persists, and this reduction reflects the specific feature set we use rather than abstraction in general---a spatial heatmap is no less an abstraction of raw gaze, yet retains more identity. The value of such a hand-designed representation is not that it competes with learned
features on raw accuracy, but that its inductive biases are interpretable and auditable: a system designer can reason in advance about what a fixed feature set can and cannot encode about identity, a transparency that end-to-end
learned representations do not readily afford. 

We make three contributions: (1)~the first systematic representation-level comparison of gaze encodings along a joint utility--privacy axis, under matched model capacity and training budget so that differences are attributable to the representation, not model size; (2)~the finding that task-preserving abstraction is not uniformly privacy-preserving---the engineered feature set retains more utility and leaks less identity than a spatial heatmap, a favorable and inspectable operating point relative to raw gaze; and (3)~a modular implementation that separates the gaze representation from the utility and re-identification probes behind a shared protocol, accommodating three structurally distinct encodings (1D streams, 2D heatmaps, and feature vectors) under one harness. The implementation will be made publicly available.

%% file: sections/rw.tex
This section situates gaze telemetry within the privacy--utility tension in
intelligent XR: its value as a context source for task-aware adaptation, the
biometric risk embedded in that same signal, and the mechanisms proposed to
reduce exposure while preserving utility.

\subsection{Utility: Gaze as an Interaction and Context Modality}
Gaze is among the most informative signals available to an XR system: it
provides a continuous behavioral proxy for attention, task state, and
near-term intent. Eye and head movement patterns differ
systematically across tasks---in fixation duration, saccade amplitude,
head-rotation velocity, and eye--head coordination---and models such as EHTask
exploit these differences to recognize user activity from gaze alone~\cite{Hu2023}.
Gaze dynamics also support predictive interaction: David-John et al.\ show that
gaze velocity and saccade structure can anticipate the onset of interaction in
VR without knowing the object being viewed~\cite{David2021}, and intent-aware
models adapt dwell-time selection to suppress false activations in gaze-only XR
input~\cite{Narkar2024, Ramiotis2025}. Aggregate gaze heatmaps can likewise
drive context-aware interface behavior such as notification placement~\cite{Ilo2024}. Across these systems the common ingredient is
\emph{fine-grained temporal structure}---the precise dynamics of fixations and
saccades on which the strongest task and intent models depend.

This is precisely what creates a privacy--utility tension. The same fine
temporal dynamics that make gaze useful for inference are also the features
most likely to encode persistent, user-specific signatures, so maximizing
task-awareness can mean preserving exactly the structure that makes users
identifiable. A key question follows: can a simplified gaze representation
retain enough task-relevant information for context-aware XR while suppressing
the biometric detail embedded in raw gaze trajectories?

\subsection{The Bio-Digital Footprint: Privacy Risks in XR Telemetry}
XR telemetry is not merely an interaction substrate; it is a behavioral
biometric record. The continuous head, hand, and gaze streams required to
stabilize rendering and support input also encode idiosyncratic patterns of how
a person looks, turns, and reaches. Nair et al.\ illustrate the scale of this
risk using head and hand motion from $55{,}541$ real VR users: after five
minutes of enrollment per person, users were identified from the full pool with
$94.33\%$ accuracy from $100$ seconds of motion and $73.20\%$ from only
ten~\cite{Nair2023}. Removing names or account identifiers is therefore
insufficient when the telemetry itself remains identifying.

Gaze intensifies this risk because oculomotor behavior is both useful for
inference and biologically distinctive, shaped by involuntary neuromuscular
dynamics that are difficult to consciously mask. Lohr and Komogortsev show that
eye-movement biometrics are relatively spoof-resistant and support continuous
authentication; their EKYT model reaches a $3.66\%$ equal error rate (EER) from only
five seconds of eye movements, with usable performance even at degraded,
VR/AR-relevant sampling rates~\cite{Lohr2022}. These signatures persist on
VR-grade hardware and across devices: using the GazeBaseVR
dataset~\cite{Lohr2023GazeBaseVR}, Aziz and Komogortsev demonstrate
cross-platform identity linkage, re-identifying users across distinct
eye-tracking platforms from their oculomotor behavior alone~\cite{Aziz2024}.

The core harm, however, is not identification within a single dataset but
\emph{linkage} across contexts. Powar and Beresford argue that dataset privacy
is best understood through linkage: risk arises whenever data can be connected
to external information that singles out or reveals attributes about a subject,
so privacy must be treated as risk management rather than the mere removal of
direct identifiers~\cite{Powar2023}. A leaked gaze stream linked to a user
through their biometric signature can then expose attention patterns, cognitive
state, or health-relevant signals; Kr\"oger et al.\ document that eye-tracking
data can reveal identity, age, gender, emotional state, personality, and
physical or mental health conditions~\cite{Kroger2020}. Because these sensitive
streams are also core to interaction and cannot simply be disabled, XR privacy
defenses must be evaluated jointly against privacy, latency, and task
utility---a constraint that motivates intervening on how gaze is
\emph{represented} rather than only on how it is later perturbed.

\subsection{Privacy Mechanisms for XR Data}
Privacy-preserving mechanisms for behavioral telemetry intervene at various
stages of the data lifecycle, from collection and streaming to model training
and release~\cite{Wilson2024}, but each faces hurdles in XR's continuous,
utility-critical sensor streams. Formal mechanisms such as Differential Privacy
(DP) and Federated Learning (FL) offer rigorous guarantees, yet DP in XR
struggles with defining the unit-of-privacy, the compounding loss of repeated
continuous releases, and steep utility costs~\cite{Ponomareva2023, Powar2023};
sample-level DP such as Kaleido can add $80$\,ms of latency, enough to induce
discomfort and render real-time XR unusable~\cite{Ibragimov2025}. FL keeps data
on-device but does not eliminate identity leakage from transmitted model
updates~\cite{Ahmed2026}.

For offline sharing, anonymization and synthetic data are common, but removing
direct identifiers is insufficient because the raw signal functions as a
persistent biometric~\cite{Liu2019, Lohr2022}: adversaries can re-link
high-dimensional traces to external profiles. Real-time perturbation---noise
injection, downsampling, smoothing---trades privacy against utility in a way
that is highly application-dependent, and mechanisms viable for casual
interaction can break the temporal precision required for competitive or
gaze-contingent use~\cite{Wilson2024, Ibragimov2025}.

Across these approaches, one design choice remains under-theorized as a privacy
mechanism in its own right: the \emph{input representation}. Raw 3D gaze
vectors, engineered statistical features, aggregated heatmaps, and learned
embeddings expose different combinations of task-relevant semantics and
identity-revealing micro-dynamics, yet the literature has not systematically
evaluated whether a task-preserving representation can itself reduce biometric
linkage while retaining utility. Because formal guarantees and heavy
perturbation often impose unacceptable latency and utility costs in interactive
XR, selecting the representation before modeling offers a promising, lightweight
alternative---stripping biometric signal while preserving the contextual
structure needed for activity recognition. We take up this question directly.

%% file: sections/threatScenario.tex
\label{sec:threat-model}
Our central question is simple: \emph{can a third party recognize the same
person from their gaze, and does the choice of gaze representation make this
easier or harder?} XR systems, like the Apple Vision Pro, rarely hand applications the raw gaze stream; they expose a derived representation. Such representations are widely assumed to be ``safe'' because they abstract the raw signal, and we treat that as a claim to
test rather than grant. The risk is a chain. If a representation lets an
adversary recognize a returning user (\emph{re-identification}), the adversary
can link that user's activity across sessions, apps, and datasets
(\emph{linkage})~\cite{Powar2023}; and a persistent, linked profile is the
basis for inferring sensitive attributes such as age, sex, cognitive state,
or health (\emph{attribute inference})~\cite{Kundu2026}. We measure the first,
enabling link in this chain and treat it as a proxy for the downstream risk.

\paragraph{System and trust boundary.}
Let $X$ be a raw gaze stream and $Z=f(X)$ a derived representation. We assume
$f$ runs inside a trusted XR layer---the operating system, platform SDK, or a
secure hardware module---and that only $Z$ crosses to third parties. The raw
stream $X$ never leaves that layer and is unavailable to third-party
applications, analytics services, remote servers, or dataset recipients.

\paragraph{Asset.}
The asset at risk is the representation $Z$ and, specifically, any
\emph{persistent biometric signature} it carries: oculomotor regularities
stable enough to recognize the same person across separate sessions. Derived
representations are routinely treated as low-risk because they abstract or
aggregate the raw signal---a heatmap, for instance, discards sample-level
dynamics~\cite{Duchowski2012,Liu2019}. Whether that abstraction actually removes
identifying information is the empirical question we test, not an assumption we
grant. Unlike a password, a leaked biometric signature cannot be reissued, so
any residual identity in $Z$ is a permanent liability~\cite{Nair2023,Lohr2022}.

\paragraph{Adversary and incentive.}
We model an \emph{honest-but-curious} third party: a recipient with legitimate
access to $Z$ but not $X$---a downstream researcher, a third-party XR
application, an analytics service, or a platform partner---who follows the
access rules but analyzes the data it is given for an unintended purpose. It
does not breach the system; it simply trains a re-identification model on the
representations it is entitled to see. The incentive is linkage: recognizing a
returning user lets the adversary stitch that user's activity together across
sessions, applications, or datasets into a profile the user never consented to.
Concretely, two applications that each receive gaze can collude to learn that
the person performing one activity in the first is the same person performing a
sensitive activity in the second; or a social-VR platform can tie a pseudonymous
attendee of a sensitive gathering back to a known identity. Such a profile is
also the substrate for attribute inference. We give the adversary
knowledge of $f$ and let it train its re-identification model from scratch,
following the strong-adversary convention in eye-movement
biometrics~\cite{DJohn2022,Lohr2022}: privacy must come from the
representation itself, not from keeping the pipeline secret.

\paragraph{Attack channels.}
We consider two. \textbf{C1 (dataset release):} a corpus of representations is
shared for reproducibility, benchmarking, or downstream modeling.
\textbf{C2 (runtime exposure):} a trusted XR layer streams $Z$ continuously to
applications, plugins, or remote services. C2 covers only data flowing
\emph{through} the system to legitimate recipients; it excludes attacks that
observe the user's eyes directly, such as a bystander or external camera, which
constitute a different threat surface.

\paragraph{What we measure.}
We quantify identity leakage as closed-set identification accuracy: each probe
is assumed to belong to one of $N$ enrolled users, and the attack succeeds when
the top-ranked prediction is the correct user. We report Top-1 accuracy against
the chance baseline $1/N$. Two clarifications, since this differs from one-to-one
authentication. First, we deliberately measure $N$-way identification: it is the standard, attacker-favorable way to express how strongly
a representation singles individuals out---its \emph{unicity} in linkage
terms~\cite{Powar2023}. Closed-set identification only requires ranking, never
rejection, so a representation that drives it toward chance is genuinely hard to
link, not merely hard to authenticate. Second, we also report Top-$k$ accuracy,
because linkage rarely requires a single exact guess; narrowing a user to a short
candidate list is often enough to act on, so Top-$k$ indexes linkability beyond
exact identification.

\paragraph{Scope.}
We measure identity leakage, not attribute inference. \emph{Attribute inference}
is the parallel threat of recovering \emph{what} a user is rather than
\emph{who} they are---demographic or state variables such as age, sex, cognitive
load, or health indicators---from the same representation~\cite{Kundu2026}. The two
are distinct: a representation can suppress identity while still leaking
attributes, so we do not claim that reducing re-identification removes all
sensitive inference. We center identity precisely because it is the enabling
step: re-identification is what lets an adversary link observations to a
persistent profile and accumulate them over time, which is what makes attribute
inference scalable and durable rather than a one-off guess~\cite{Kundu2026}. We
exclude threats that are orthogonal to representation choice and addressed
elsewhere---OS or driver compromise, physical side channels, network-traffic
analysis, social engineering, and multimodal linkage with RGB, audio, or hand
telemetry~\cite{Nair2023, Lohr2022}. Our contribution is the gap these leave open: how the \emph{choice of gaze representation}---not the raw
data, and not a downstream noise mechanism---governs identity leakage.

%% file: sections/methods_revised.tex
This section describes how we compare three gaze representations under
two competing objectives: \emph{utility}, operationalized as
coarse-grained action recognition, and \emph{privacy risk},
operationalized as closed-set user re-identification. We begin with the
computing environment and software stack, then introduce the HoloAssist
dataset and the synchronized eye-gaze and head-pose streams it provides.
From this shared signal we derive the three representations under
study---raw gaze, spatial attention heatmaps, and engineered oculomotor
features---and detail the preprocessing for each. We then describe the
encoder paired with every representation and the two-phase training
protocol applied uniformly across them. By holding the dataset, task
definitions, data splits, and training procedure fixed, the design
isolates representation as the experimental factor of interest, so that
differences in utility or privacy can be attributed to representational
choice rather than to incidental modeling decisions.

\subsection{Hardware \& Software}
\label{subsec:methods-hardware}

All experiments were executed on a GPU cluster at our university's computing center. Jobs were dispatched
through the Slurm workload manager to the cluster's GPU partition;
following the recommended GPU-to-CPU binding, each job requested a
single NVIDIA L40S GPU (48~GB) with its topologically affiliated CPU
cores and system memory.
Models were implemented in PyTorch~2.12.0 (CUDA~12.8 build) with Python~3.12. Our code, the complete software environment, and all training configurations will be made publicly available as a repository.

\subsection{HoloAssist Dataset}
\label{subsec:methods-dataset}

We selected HoloAssist because it stresses both of our research
questions far more than the datasets typically used for this work.
HoloAssist is a large-scale egocentric dataset in which a task performer
completes real, physical manipulation tasks while wearing a
mixed-reality headset, with a remote instructor guiding them; the
headset captures seven synchronized streams, including the eye-gaze and
head-pose signals we use~\cite{HoloAssist}. This setting is more
naturalistic, and therefore more challenging, than the scripted
viewing tasks common in prior gaze-based work.

The most consequential difference is one of scale. Because the test
labels are withheld, we draw all of our statistics from the combined
training and validation data. As released, this material contains
$12{,}542$ coarse-grained action events across $1{,}758$ sessions,
spanning $39$ distinct verbs, $89$ nouns, and $403$ verb--noun
combinations (for example, a verb such as \emph{insert} paired with a noun such as \emph{battery} forms the action class \emph{insert battery}), recorded from more than $200$ distinct identities. For our
experiments we use a filtered subset of these events (the rationale and
criterion are given in Section~\ref{subsec:methods-preprocessing}),
yielding $10{,}212$ events across $1{,}754$ sessions over $21$ verbs,
$44$ nouns, and $85$ verb--noun action classes; the identity pool is
unaffected. Either way, the scale dwarfs that of comparable gaze-task
datasets, which recruit on the order of tens of participants---EHTask,
for example, used $30$~\cite{Hu2023}. This gap matters for both of our
tasks. For action recognition, an $85$-way verb--noun label space is far
larger and more realistic than the handful of categories used in prior
studies, so the task is correspondingly harder. For re-identification,
the difficulty of the problem grows with the number of candidate
identities: distinguishing one person among more than $200$ is much
harder than among $30$, and the accuracy expected from random guessing is
correspondingly lower. HoloAssist therefore lets us evaluate the
privacy--utility tradeoff on a decision space that is demanding along both
axes at once, which is exactly the regime in which representation choice
is most likely to matter.

\subsection{Data Preprocessing}
\label{subsec:methods-preprocessing}

To study how the level of abstraction in a gaze signal affects both task
recognition (utility) and user identification (privacy risk), the raw
synchronized streams of gaze-direction vectors and head-pose quaternions
were transformed into three representations.

Before this transformation, we filtered the action label space to remove
classes too rare to be learned or evaluated reliably. We retained only
verb--noun action classes with at least $30$ labeled events and
appearances in at least $3$ distinct recording sessions; classes below
either threshold were dropped. This reduces the action vocabulary from
$403$ to $85$ classes (and the event count from $12{,}542$ to $10{,}212$,
about an $18\%$ reduction) while removing only long-tail classes for
which per-class accuracy would be dominated by sampling noise. The
session-count threshold additionally ensures that no class is supported
by a single recording, so that recognition cannot be driven by
session-specific artifacts.

\paragraph{Raw gaze (EHTask channels).}
The least-processed representation preserves the eye and head movements
themselves as time series. Following the EHTask formulation~\cite{Hu2023},
the HoloAssist signals are mapped into three two-channel streams:
Eye-in-Head (EiH), obtained by rotating the world gaze by the inverse
head quaternion; head rotational velocity, derived from ZYX Euler angles;
and Gaze-in-World (GiW)~\cite{diaz2013real}.

\paragraph{Spatial attention heatmaps.}
This representation captures \emph{where} a person looked rather than the
moment-to-moment motion of their eyes. Gaze and head pose are combined
into a single global look direction, which is projected onto a
two-dimensional equirectangular grid spanning the full field of regard
($360^{\circ}$ horizontal by $180^{\circ}$ vertical). Accumulating these
projected gaze points over short temporal snapshots, with optional
Gaussian smoothing, yields an image-like heatmap---brightest where the
user looked most---centered on the user's head orientation.

\paragraph{Engineered oculomotor features (statistical).}
This representation replaces the raw gaze trajectory with a compact, fixed
vector of interpretable summary statistics describing how the eyes behaved
within each analysis window. The raw stream is segmented into sliding
windows, and for each window we compute the $55$-dimensional continuous
oculomotor feature set of Narkar et al.~\cite{Narkar2024}---window-level
descriptors of gaze velocity, acceleration, position, and dispersion---%
building on engineered gaze-feature representations used for privacy
analysis by David-John et al.~\cite{DJohn2022} and grounded in the
eye-movement feature taxonomy of Rigas et al.~\cite{Rigas2018}. We compute
these features directly from the windowed signal, without an explicit
fixation/saccade event-detection stage, so the representation does not
depend on event-segmentation thresholds. The resulting feature vectors are
$z$-score normalized using a scaler fit exclusively on the training split;
the full feature list is provided with the code release.

\paragraph{Why these representations.}
The three representations span a deliberate spectrum from minimally
processed to heavily abstracted. Raw gaze retains the signal in nearly
its original form, including the fine, person-specific motor patterns
that make eye movements behave like a biometric. Heatmaps keep \emph{where}
a person looked but discard the dynamics of \emph{how} the eyes moved to
get there. Engineered features reduce the signal further, to a handful of
summary statistics. Intuitively, moving along this spectrum should trade
descriptive detail for privacy: the more a representation abstracts away
the raw motor signal, the less identifying information it can expose---%
though whether it also sheds information the task needs is exactly what
our experiments test. These expectations frame the two hypotheses we
evaluate in the Results section, both posed as raw-versus-alternative
contrasts: relative to raw gaze, each alternative representation
preserves action-recognition utility (H1), while substantially reducing
re-identification accuracy (H2).

\begin{figure}[t]
  \centering
  \includegraphics[width=0.92\linewidth]{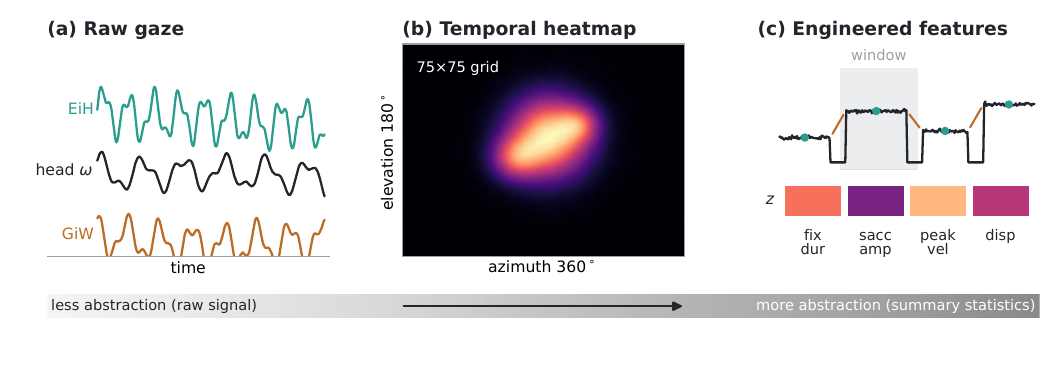}
  \caption{The three gaze representations, from the same eye and head
  stream: (left) raw gaze/head signal, (center) spatial attention heatmap,
  (right) engineered continuous-feature vector. Abstraction increases left
  to right, discarding fine-grained temporal structure that carries identity.}
  \label{fig:representations}
\end{figure}

\paragraph{Why engineered features should attenuate identity.}
The engineered representation occupies the most abstracted end of this
spectrum (Figure~\ref{fig:representations}, right), and the \emph{form} of
its abstraction is what we expect to drive the privacy result. Eye-movement
biometrics draw their discriminative power from fine, person-specific
temporal microstructure---the precise kinematics of individual saccades and
the moment-to-moment stability of fixations---which prior work shows can
identify users from only seconds of gaze~\cite{Lohr2022, DJohn2022}. The
continuous feature set summarizes each window with distributional
descriptors of velocity, position, and dispersion~\cite{Narkar2024,
Rigas2018}; this averaging discards the within-window ordering and fine
timing that carry identity, while retaining the coarser regularities that
separate one task from another. The engineered representation should
therefore attenuate the biometric signal \emph{by construction}---through
the choice of what to encode, rather than through added noise or a learned
adversary. This is the literature-grounded basis for our two hypotheses
(Section~\ref{sec:hypotheses}): that abstraction along this gradient trades
identity for utility \emph{asymmetrically}. Whether a representation that
abstracts \emph{spatially} (the heatmap) behaves the same way is an open
question our experiments address.

\subsection{Model Architectures}
\label{subsec:methods-architectures}

\begin{figure*}[t]
  \centering
  \includegraphics[width=0.85\textwidth]{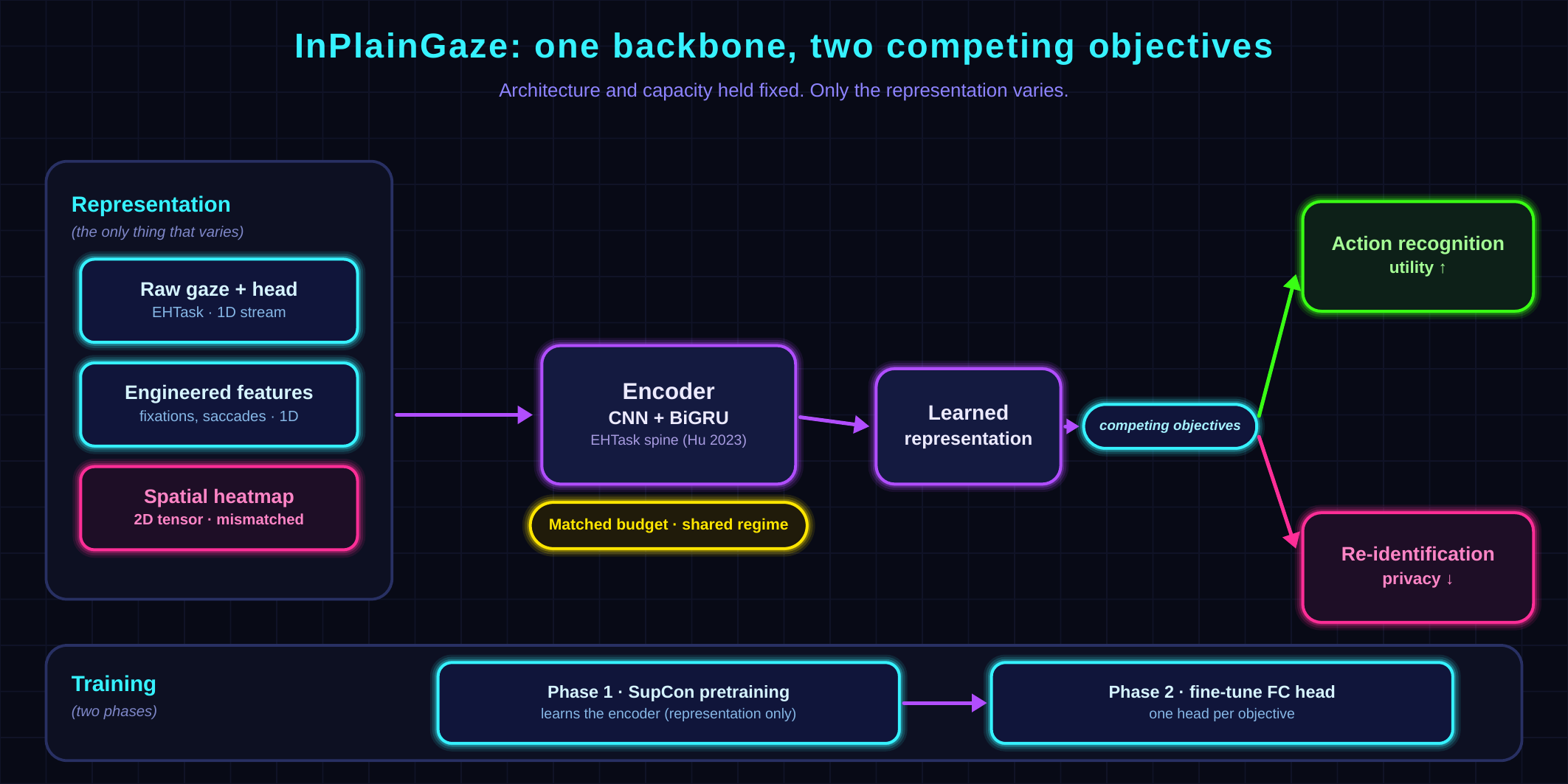}
  \caption{Comparison setup. Only the input gaze representation varies; the
  CNN+BiGRU encoder, its capacity, and the training budget are fixed. The
  learned representation is read out on two competing objectives---action
  recognition (utility) and closed-set re-identification (privacy)---after
  two-phase SupCon pretraining and per-objective fine-tuning.}
  \label{fig:setup}
\end{figure*}

Because the study compares representations rather than networks, each
representation is paired with an encoder drawn from a common family, and
the encoders are matched as closely as the input format permits, so that
performance differences reflect the representation and not the model
(Figure~\ref{fig:setup}). All
three encoders share the same backbone: one or more convolutional layers
that extract local patterns, followed by a bidirectional gated recurrent
unit (BiGRU) that integrates those patterns over time. The representations
differ only in input shape, which dictates the sole necessary
architectural difference. Raw gaze is a multi-channel one-dimensional
time series and is encoded with a 1D~CNN and GRU matching the EHTask
reference architecture~\cite{Hu2023}; the engineered features form a
per-window feature sequence and are likewise encoded with a
1D~CNN~+~BiGRU; the heatmaps are image-like 2D maps per time step, so a
2D~CNN replaces the 1D convolution to accommodate the spatial input
before the same recurrent stage.

To ensure that observed differences reflect the representation rather
than incidental modeling choices, the three encoders are harmonized along
the dimensions that would otherwise confound the comparison. Normalization
is applied consistently across branches---batch normalization is used in
the convolutional and fully connected stages of all three encoders, not
only the 1D streams---and the depth of the classifier head is matched
across representations. All three pipelines are trained under an identical
compute budget and a shared hyperparameter configuration. The encoders
consequently differ only in the input-driven distinction described
above---a 2D convolution for the spatial heatmap in place of the 1D
convolution used for the temporal streams---rather than in capacity,
normalization, or training budget.

\subsection{Training}
\label{subsec:methods-training}

Every representation is trained with the same two-phase protocol, so that
the training procedure is held constant across the comparison. In the
first phase, the encoder backbone is pretrained with a supervised
contrastive objective (SupCon)~\cite{Khosla2020} and no classification
head: the network learns an embedding space in which examples that share
the relevant label are pulled together and all others are pushed apart.
The grouping label is matched to the downstream task---coarse-grained
action class for the action-recognition pipeline, and performer identity
for the re-identification pipeline---so that the embedding is organized
around the quantity each task must predict. In the second phase, the
pretrained backbone is frozen, a freshly initialized fully connected
classification head is appended, and only this head is trained, using
cross-entropy loss with the AdamW optimizer and an exponential
learning-rate schedule. The re-identification classifier uses early
stopping on validation loss (patience $5$); the pretraining and
action-classification phases run for a fixed epoch budget. Full
hyperparameters for all three phases---per-representation layer sizes,
learning-rate schedules, batch size, and epoch budgets---are listed with the
code release.

%% file: sections/results_revised.tex
Eye and head motion are useful because they are informative, and that
same informativeness is part of what can make them identifying. We take
the \emph{raw} gaze representation as our reference point and ask how each
alternative representation compares to it along two axes: how well it
supports coarse-grained action (CGA) classification, and how much it
limits an attacker's ability to re-identify a user. We first fix the
evaluation protocol and state the two hypotheses, then give a high-level
overview of what the result tables show at a glance, and then report
the utility axis and the privacy axis in detail.

\subsection{Evaluation Metrics and Comparisons}
\label{sec:metrics}

\begin{figure}[t]
\centering
\includegraphics[width=0.82\linewidth]{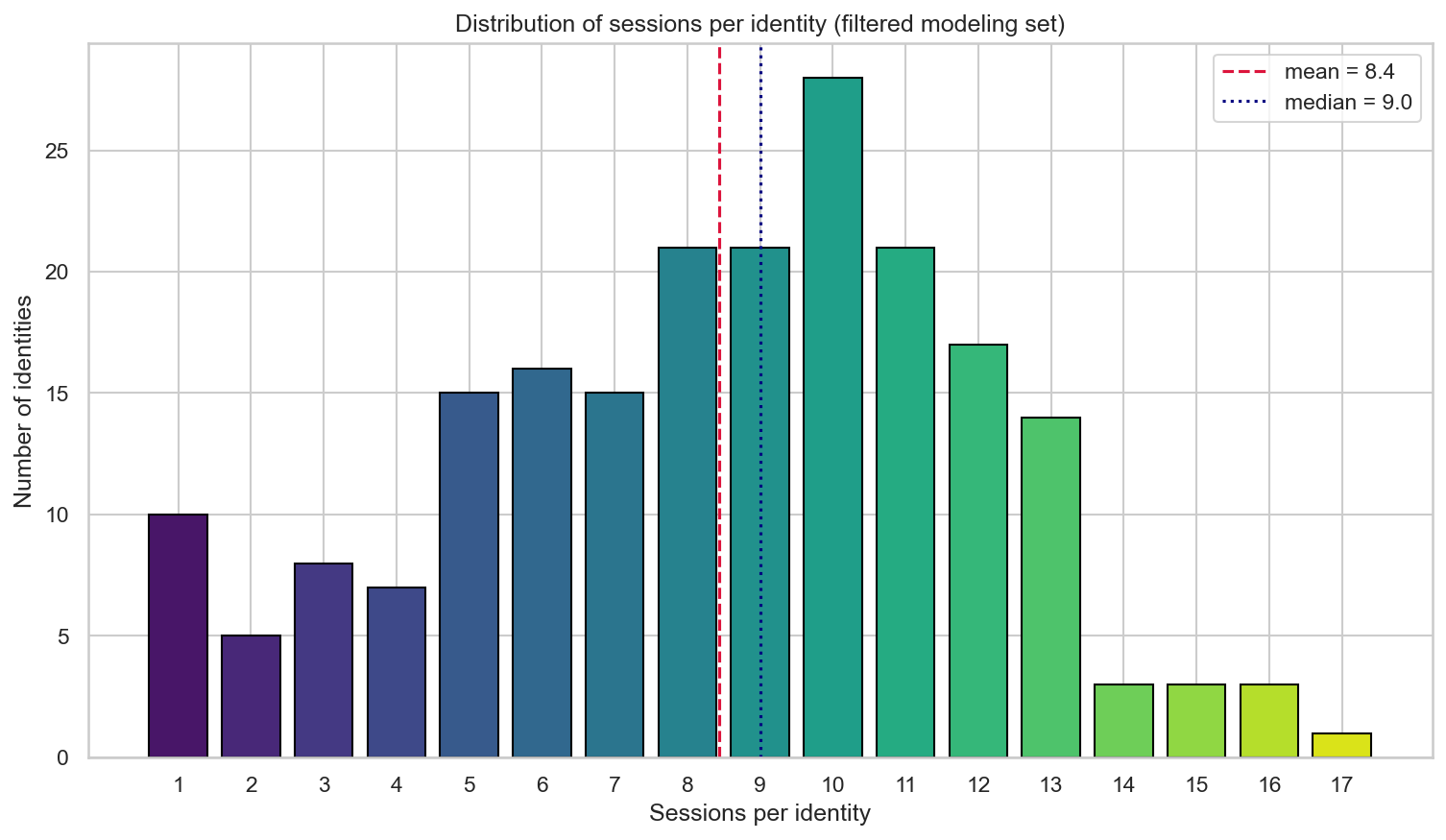}
\caption{Recording sessions per identity in the filtered HoloAssist subset.
Most identities contribute several sessions, so the stratified-by-identity
split holds out segments while keeping every identity represented across
train, validation, and test.}
\label{fig:sessions}
\end{figure}

We use a stratified 80/10/10 split at the coarse-grained action (CGA)
segment level, stratified by participant identity, following the
closed-set identification protocol of David-John et al.~\cite{DJohn2022}.
All $N{=}206$ identities appear in the training, validation, and test
splits; the units that are held out are CGA segments rather than whole
sessions, so segments from the same session can appear in different
splits. Action recognition is therefore measured in a within-identity,
cross-segment setting, which separates the effect of the representation
from the effect of which identities are present. The same split defines
the attacker for the privacy axis: it represents a permissive attacker
who already has enrollment data for every candidate identity, recorded in
the same context (Figure~\ref{fig:sessions}). The re-identification numbers reported below are
therefore upper bounds under this enrollment-rich setting.
Because the two tasks call for different measures, we report
task-appropriate metric sets. For action recognition we report
\emph{top-1 accuracy} (primary), \emph{top-5 accuracy}, and
\emph{balanced accuracy}---the mean of per-class recalls, which prevents
frequent classes from dominating the score on the imbalanced $85$-way
label space. For re-identification we report \emph{top-1 accuracy}
(primary), \emph{top-5 accuracy}, and \emph{mAP@R} (mean average
precision at $R$), a retrieval-style score that rewards ranking
same-identity examples ahead of others, as defined by Musgrave et
al.~\cite{Musgrave2020}. To account for variation due to random seeding,
each (representation $\times$ task) configuration is trained and evaluated
independently under three pre-committed seeds,
$\mathcal{S}=\{55, 777, 1337\}$. We designate seed $55$ as the primary
seed and report seeds $777$ and $1337$ as robustness checks; tables give
the seed-level mean $\pm$ standard deviation across all three. (An
additional re-identification run on a separate tuning seed, used only for
hyperparameter selection, is excluded from all reported results.)
Utility (H1) is reported as effect sizes alongside the pre-registered
equivalence test; privacy (H2) is reported as the reduction in
re-identification accuracy relative to raw gaze.

\subsection{Hypotheses}
\label{sec:hypotheses}
We compare each alternative gaze representation (spatial heatmap,
engineered oculomotor features) against the raw gaze representation along
two axes: utility for coarse-grained action recognition, and resistance
to closed-set re-identification.

\begin{enumerate}
  \item \textbf{(Utility)} Relative to raw gaze, each alternative
  representation incurs only a modest reduction in coarse-grained
  action-recognition accuracy. We quantify the size of this reduction
  directly, so that the utility cost of a representation can be weighed
  against the privacy gain it provides (H2) rather than reduced to a
  single equivalence verdict; for transparency we also report the
  pre-registered equivalence test (TOST, $\Delta=0.05$)~\cite{lakens2018}.
  \item \textbf{(Privacy)} Relative to raw gaze, each alternative
  representation significantly reduces closed-set re-identification
  accuracy. We judge ``significant'' by the size of the reduction and its
  consistency across seeds rather than by a fixed multiple.
\end{enumerate}

\subsection{Results Overview}
\label{sec:results_overview}

Before any statistical testing, two patterns stand out, and they map onto
our two hypotheses. On the utility axis (action recognition,
Figure~\ref{fig:utility}, higher is better), raw gaze achieved the highest
accuracy in every seed, with the engineered representation close behind and
the heatmap further back---the same ordering on top-1, top-5, and balanced
accuracy. On the privacy axis (re-identification, Figure~\ref{fig:reid},
lower is better), the three representations separate far more, and the
engineered representation is the most private on every metric---top-1
$0.0190$, top-5 $0.0795$, mAP@R $0.0036$, against $0.1917$ / $0.5198$ /
$0.0596$ for raw and $0.1302$ / $0.4112$ / $0.0223$ for the heatmap. The
best representation therefore differs by objective: raw gaze is most
accurate, the engineered representation most private.

\subsection{Utility: Action Classification (H1)}
\label{sec:results_h1}

This axis quantifies how much action-recognition accuracy each
alternative representation gives up relative to raw gaze. Rather than
collapse this into a single pass/fail equivalence verdict, we report the
size of the gap directly; the Discussion then weighs it against the
privacy gain from Section~\ref{sec:results_h2}. We summarize the gap as
$\hat{\delta}=\mathrm{Acc}_{\text{Raw}}-\mathrm{Acc}_{\text{Alt}}$ in
top-1 accuracy (positive values mean raw scored higher), aggregated
across the three seeds as mean and across-seed standard deviation, on the
primary $85$-way verb--noun action task. For transparency we also report
the pre-registered equivalence test (TOST, $\Delta=0.05$, $90\%$ CI).
Seeds are treated as the replication unit ($n=3$); we flag the limited
seed count as a constraint on statistical power.

\paragraph{Raw vs.\ Engineered.}
On the primary action task, the engineered representation trailed raw
gaze by $\hat{\delta}=0.033$ in top-1 accuracy (per-seed $0.044$, $0.022$,
$0.034$; $90\%$ CI $[0.014, 0.052]$), a relative reduction of roughly
$15\%$ from raw's $0.216$. The pre-registered equivalence test is
inconclusive at $\Delta=0.05$: the interval's upper bound ($0.052$) lies
just outside the equivalence bound, so the test establishes neither
equivalence nor a meaningful difference. The relative gap is smaller on
coarser labels: for verb-level recognition ($21$ classes) the engineered
representation trails raw by $\hat{\delta}=0.040$ on a larger base---about
a $9.5\%$ relative reduction---with a tighter interval ($[0.031, 0.049]$)
that does satisfy equivalence at $\Delta=0.05$.

\paragraph{Raw vs.\ Heatmap.}
The heatmap gives up considerably more utility, and far less stably: it
trailed raw by $\hat{\delta}=0.113$ on average (per-seed $0.170$, $0.116$,
$0.052$), with an across-seed standard deviation roughly an order of
magnitude larger than the engineered representation's, and no seed
approaching the $\Delta=0.05$ equivalence bound.

\begin{figure}[t]
\centering
\includegraphics[width=0.92\linewidth]{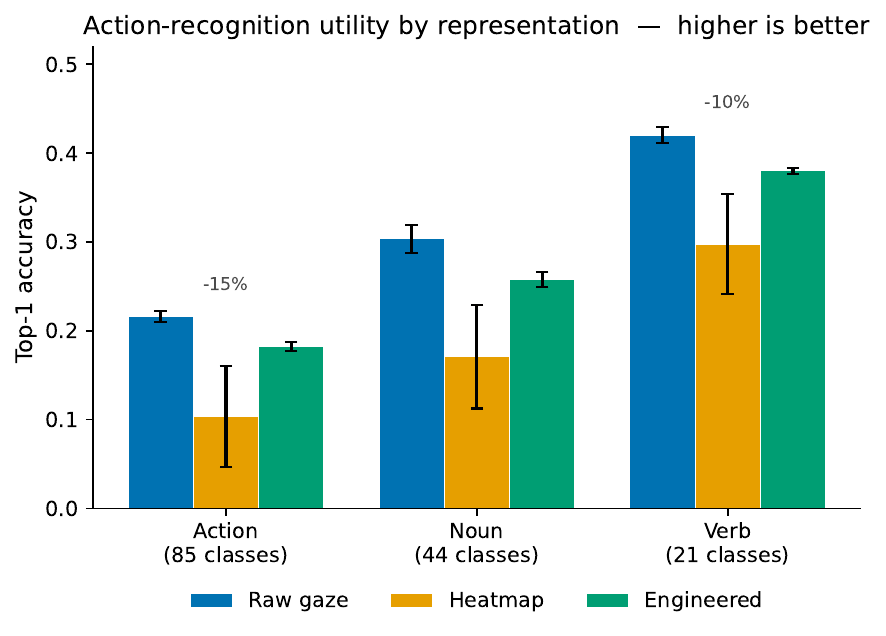}
\caption{Action-recognition accuracy across representations and label
granularities (higher is better). The engineered representation tracks raw
gaze closely, with the gap narrowing on the coarser verb-level task.}
\label{fig:utility}
\end{figure}

Taken together, the engineered representation gives up about $15\%$ of raw
gaze's top-1 action accuracy---less on coarser labels---while the heatmap
gives up substantially more (Figure~\ref{fig:utility}).
Section~\ref{sec:results_h2} quantifies the privacy gain that this utility
cost buys.

\subsection{Privacy: Re-Identification (H2)}
\label{sec:results_h2}

\begin{figure}[t]
\centering
\includegraphics[width=0.92\linewidth]{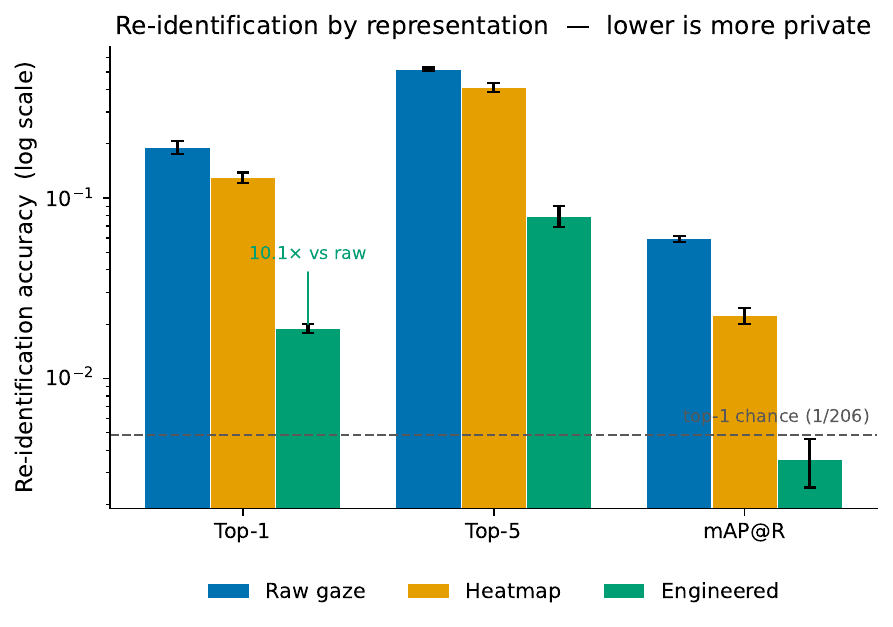}
\caption{Closed-set re-identification across representations (log scale;
lower is more private). The engineered representation reduces top-1
re-identification by roughly an order of magnitude relative to raw gaze
while remaining above the $1/206$ chance floor.}
\label{fig:reid}
\end{figure}

This axis measures how much each alternative representation lowers
re-identification accuracy relative to raw gaze; lower values indicate
stronger privacy (Figure~\ref{fig:reid}). The chance top-1 rate is
$\approx 1/206 = 0.49\%$. For each representation we report the size of the
reduction relative to raw gaze and its consistency across the three seeds.

\paragraph{Engineered vs.\ Raw.}
Top-1 re-identification fell from $0.1917$ (raw) to $0.0190$ (engineered),
a reduction of $10.1\times$ (Figure~\ref{fig:reid}), and mAP@R fell from
$0.0596$ to $0.0036$ ($16.6\times$); top-5 fell from $0.5198$ to $0.0795$
($6.5\times$). This roughly order-of-magnitude top-1 reduction is
consistent across all three seeds, supporting H2 for the engineered
representation. The engineered top-1 value of $0.0190$ remains above the
chance rate, by a factor of about $3.9$.

\paragraph{Heatmap vs.\ Raw.}
Top-1 re-identification fell from $0.1917$ (raw) to $0.1302$ (heatmap), a
reduction of $1.47\times$, and mAP@R fell from $0.0596$ to $0.0223$
($2.67\times$). The heatmap's top-1 reduction is far smaller---well under
an order of magnitude---so H2 is supported for the engineered
representation but not for the heatmap.

\begin{figure}[t]
\centering
\includegraphics[width=0.85\linewidth]{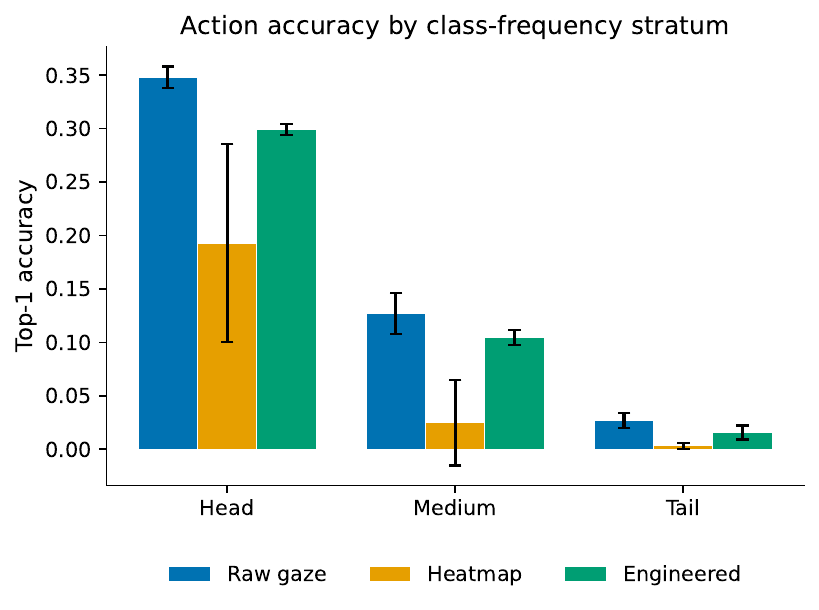}
\caption{Action top-1 accuracy by class-frequency stratum. All
representations degrade sharply on tail (rare) classes, which accounts for
the low balanced-accuracy
values on the primary task.}
\label{fig:longtail}
\end{figure}

%% file: sections/discussion_revised.tex
On the utility axis, the engineered representation did not reach
statistical equivalence with raw gaze under our pre-registered TOST. We
are careful, however, not to read this as evidence of a meaningful
difference: equivalence testing is asymmetric, and failing to establish
equivalence within a chosen bound is not the same as demonstrating a
practically important gap~\cite{lakens2018}. The observed difference is
small in absolute terms---roughly $3.3$ percentage points of top-1
accuracy on the $85$-way action task, about a $15\%$ relative reduction,
narrowing to under $10\%$ on coarser verb-level labels---and our
equivalence test was inconclusive rather than negative: its confidence
interval fell just outside the equivalence bound, establishing neither
equivalence nor a meaningful difference. The appropriate reading is
therefore that the engineered representation recovers most of the task
signal carried by the raw stream, at a modest and quantified cost, rather
than matching it outright.

Two methodological caveats temper the H1 outcome. First, our SESOI of
$5$~pp was specified a priori from related work rather than calibrated
against observed model behavior; a bound grounded in run-to-run variance,
or in a minimally noticeable difference for a downstream XR application,
would be more domain-appropriate and remains a defensible subjective
justification~\cite{lakens2018}.
Second, HoloAssist's long-tailed action distribution depresses absolute
accuracy across all three representations and likely compresses the
resolvable effect size: a small number of head classes dominate the
signal while rare tail classes contribute disproportionately to error
(Figure~\ref{fig:longtail}). Notably, our gaze-only models reach this
accuracy range without the RGB and hand-pose modalities that the
HoloAssist authors required to push a multimodal TimeSformer baseline to
roughly $50\%$ top-1 on coarse-grained actions~\cite{HoloAssist}. We do
not read this gap as a ceiling on gaze: the transformer baselines differ
from our models along three axes at once. They are substantially
larger-capacity architectures; they require correspondingly large
training corpora, which gaze-only XR datasets of this kind do not yet
provide; and they consume RGB video and hand pose, whereas our models see
only eye gaze and head motion. A like-for-like comparison would hold these
constant, and we read the gap as motivating---rather than foreclosing---%
gaze-specific architectures and privacy-preserving multi-stream fusion as
future work (Section~\ref{sec:future}). That a lightweight, gaze-only
representation recovers usable task signal under these constraints is
consistent with our broader claim that representation choice---not added
sensing or added noise---can be the operative design variable.

Under our pre-registered privacy criterion, the picture is asymmetric:
relative to raw gaze, the engineered representation reduces closed-set
re-identification by roughly an order of magnitude, satisfying H2, whereas
the heatmap's reduction is far smaller and does not. The strongest privacy
thus came from the engineered representation rather than the heatmap---an
inversion of our initial expectation, since the heatmap's spatial
construction might have seemed the more naturally privacy-preserving of
the two. The engineered representation retained roughly $85$--$90\%$ of
the raw stream's task signal while substantially attenuating identity
leakage relative to raw, the most favorable privacy--utility position we
observed. The substantive question motivating this work---can a gaze
representation preserve ecological task signal while attenuating the
biometric channel encoded in oculomotor dynamics?---is therefore answered
affirmatively. To our knowledge, this is the first systematic
representation-level comparison of this tradeoff for continuous gaze
action recognition in XR. It complements, rather than displaces, prior
data- and feature-level protections~\cite{DJohn2022, Liu2019}, and
establishes representation choice as a first-class privacy lever in XR
system design.

\begin{figure}[t]
\centering
\includegraphics[width=0.7\linewidth]{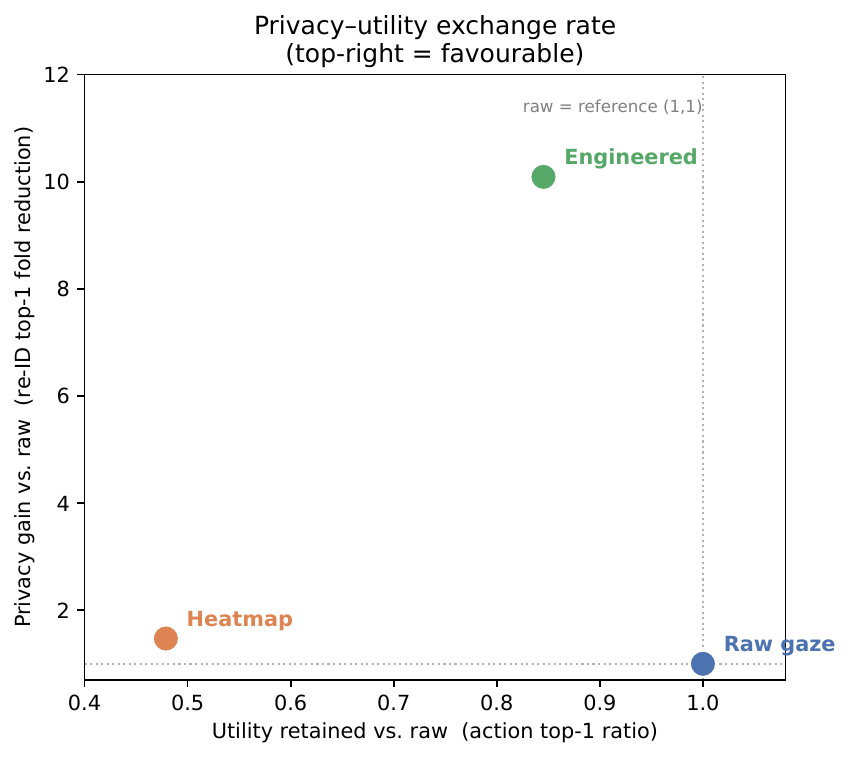}
\caption{The privacy--utility exchange rate: utility retained vs.\ raw gaze
(action top-1 ratio) against privacy gain vs.\ raw gaze (re-identification
top-1 fold reduction); raw sits at $(1,1)$. The engineered representation
occupies the favorable upper-right region.}
\label{fig:exchange}
\end{figure}

Read together, the two axes give the result that motivates this work.
Relative to raw gaze, the engineered representation trades a modest
utility cost for a large privacy gain---an exchange that places it in a
region of the design space the raw signal cannot reach: most of the
utility, an order of magnitude less identity leakage
(Figure~\ref{fig:exchange}). Crucially, this lever is lightweight. The
engineered representation is a fixed, deterministic transform applied once
at feature extraction; unlike data-level mechanisms such as differential
privacy, it spends no privacy budget, adds no calibrated noise, and
introduces no accuracy-degrading perturbation at inference time. The
privacy--utility tradeoff is therefore not a fixed tax paid for any
abstraction---the choice of representation moves the operating point, and
chosen well, it moves it in the direction a privacy-conscious system would
want. This engineered-versus-raw contrast is the robust core of our
findings and does not depend on how any other representation is tuned.

The heatmap's underperformance warrants careful interpretation. Its
representational argument---that a spatial density grid cannot, by
construction, encode saccade kinematics, fixation stability, or oculomotor
timing---remains sound, and is exactly the structural privacy property
that data-level perturbation does not provide. The result is thus a
constrained test of the heatmap, not a refutation: all three branches
share an EHTask-derived CNN+BiGRU backbone~\cite{Hu2023} built for 1D
streams and structurally mismatched to the heatmap's 2D grid (even after
we harmonized normalization and head depth), and the matched-budget regime
that keeps representation the only variable may leave the
higher-dimensional heatmap branch undertuned relative to a
representation-tailored search.

Our findings are bounded by four factors, three discussed above: the
1D-oriented EHTask backbone structurally disadvantages the 2D heatmap, so
the comparison is fair across representations but not architecture-optimal;
the shared matched-budget regime isolates representation as the variable
but may leave higher-dimensional representations undertuned; and the
equivalence bound was set a priori rather than calibrated to our data. The
fourth is reliance on a single corpus: HoloAssist's long-tailed verb
distribution interacts with representation in ways single-dataset
evaluation cannot disentangle, leaving cross-dataset replication an open
priority.

\label{sec:future}
Several directions follow from these limitations. A
representation-appropriate backbone study---pairing the heatmap with a 3D
CNN or spatiotemporal transformer---would clarify whether its
underperformance here reflects the representation or the matched-budget
architectural constraint, and is the most direct test of the
constructional privacy argument. A second avenue is joint
privacy--utility optimization: our two-stage SupCon
pipeline~\cite{Khosla2020} decouples representation learning from
classification, but an end-to-end objective that simultaneously maximizes
task utility and minimizes an adversarial identity head---drawing on
disentanglement approaches recently applied to
gaze~\cite{Elfares2025}---could yield representations Pareto-superior to
any studied here. Third, such architectures will require larger and more
ecologically diverse gaze corpora than current action-recognition datasets
provide; egocentric capture platforms such as Meta's Project
Aria~\cite{AriaGen2} offer a path toward them. Our threat model also
includes a runtime-exposure channel (C2) that we do not exercise
experimentally; evaluating representation-level defenses against an
adversary that reads $Z$ from a live XR stream---rather than from a
released corpus---is a natural next step, and would connect these results
to online, latency-bound deployment.
Finally, a representation-level privacy claim is incomplete without
auditing residual leakage across co-recorded XR sensor streams: head and
body motion alone have been shown to re-identify users at rates comparable
to gaze~\cite{Nair2023}, so cross-stream privacy auditing is a necessary
companion to representation-level protections.

%% file: sections/conclusion_revised.tex
This work positioned the \emph{representation} of gaze telemetry as a
privacy lever in XR---one exercised at feature extraction, before any
data-level mechanism. A continuous engineered feature set retained the
bulk of action-classification utility at a modest top-1 cost relative to
raw gaze, while cutting closed-set re-identification to roughly four times
chance across $N{=}206$ identities.

The value of a hand-designed representation here is not that it
outperforms learned features, but that its inductive biases are
inspectable: a designer can reason about what a fixed encoding can and
cannot reveal about identity---a lever that complements data-level
mechanisms like differential privacy rather than replacing them.